\pdfoutput=1

\documentclass[11pt]{article}

\usepackage{acl}

\usepackage{times}
\usepackage{latexsym}
\usepackage{algorithm}
\usepackage{algorithmic}
\usepackage{amsmath}
\usepackage[T1]{fontenc}
\usepackage{graphicx}
\usepackage[utf8]{inputenc}
\usepackage{microtype}
\usepackage{multirow}
\usepackage[normalem]{ulem}
\usepackage{booktabs}
\useunder{\uline}{\ul}{}
\usepackage{times}
\usepackage{bbding}
\usepackage {adjustbox}
\usepackage{amsmath,amsfonts,amssymb,amsthm, bm}
\usepackage[T1]{fontenc}

\usepackage[utf8]{inputenc}
\usepackage{graphicx}
\usepackage{microtype}
\usepackage{colortbl} 
\usepackage{inconsolata}
\usepackage[T1]{fontenc}

\usepackage[utf8]{inputenc}

\usepackage{microtype}

\title{GenDec: A robust generative Question-decomposition method for Multi-hop reasoning}

\author{Jian Wu$^{1}$, Linyi Yang $^{2}$, Yuliang Ji, Wenhao Huang$^{3}$ \\
\bf Börje F. Karlsson$^{3}$, Manabu Okumura$^{1}$ \\
$^{1}$Tokyo Institute of Technology \quad
$^{2}$Westlake University \\
$^{3}$Beijing Academy of Artificial Intelligence \\
}
  
\begin{document}
\maketitle
\begin{abstract}
Multi-hop QA (MHQA) involves step-by-step reasoning to answer complex questions and find multiple relevant supporting facts. 
However, Existing large language models'(LLMs) reasoning ability in multi-hop question answering remains exploration, which is inadequate in answering multi-hop questions. Moreover, it is unclear whether LLMs follow a desired reasoning chain to reach the right final answer.
In this paper, we propose a \textbf{gen}erative question \textbf{dec}omposition method (GenDec) from the perspective of explainable QA by generating independent and complete sub-questions based on incorporating additional extracted evidence for enhancing LLMs' reasoning ability in RAG.
To demonstrate the impact, generalization and robustness of Gendec, we conduct two experiments, the first is combining GenDec with small QA systems on paragraph retrieval and QA tasks. We secondly examine the reasoning capabilities of various state-of-the-art LLMs including GPT-4 and GPT-3.5 combined with GenDec.
We experiment on the HotpotQA, 2WikihopMultiHopQA, MuSiQue, and PokeMQA datasets. 
\end{abstract}

\section{Introduction}

In the field of natural language processing, Multi-hop Question Answering (MHQA) tasks entail iterative reasoning across diverse informational sources, such as text paragraphs. Recent advancements have demonstrated that Large Language Models (LLMs) can achieve performance comparable to that of models fine-tuned for this specific task.

Retrieval-augmented generation (RAG) represents a significant enhancement to LLMs by incorporating relevant knowledge retrieval, thereby showing considerable promise in reducing LLM-generated hallucinations and improving the overall quality of responses. This, in turn, promotes the broader application of LLMs in practical scenarios \cite{Tang2024MultiHopRAGBR}. Nonetheless, the core aspect of LLMs' reasoning capability—achieving correct answers through accurate reasoning chains—remains under investigation, particularly regarding its potential to further augment LLMs' performance.

Moreover, \citet{tang2020multi} introduced a dataset of human-verified sub-questions derived from HotpotQA~\cite{yang-etal-2018-hotpotqa} and conducted experiments focused on sub-question reasoning. Their findings reveal that models such as DFGN\cite{qiu-etal-2019-dynamically}, DecompRC\cite{min2019multi}, and CogQA\cite{ding2019cognitive}, while capable of correctly answering the overarching multi-hop questions, exhibit significant deficiencies in addressing sub-questions. This underscores a prevalent issue wherein models may circumvent the necessary reasoning process, thus failing to deduce intermediate answers to sub-questions, highlighting a critical area for further research and development in MHQA systems.

\begin{figure*}
\centering
\includegraphics[width=1.0\textwidth]{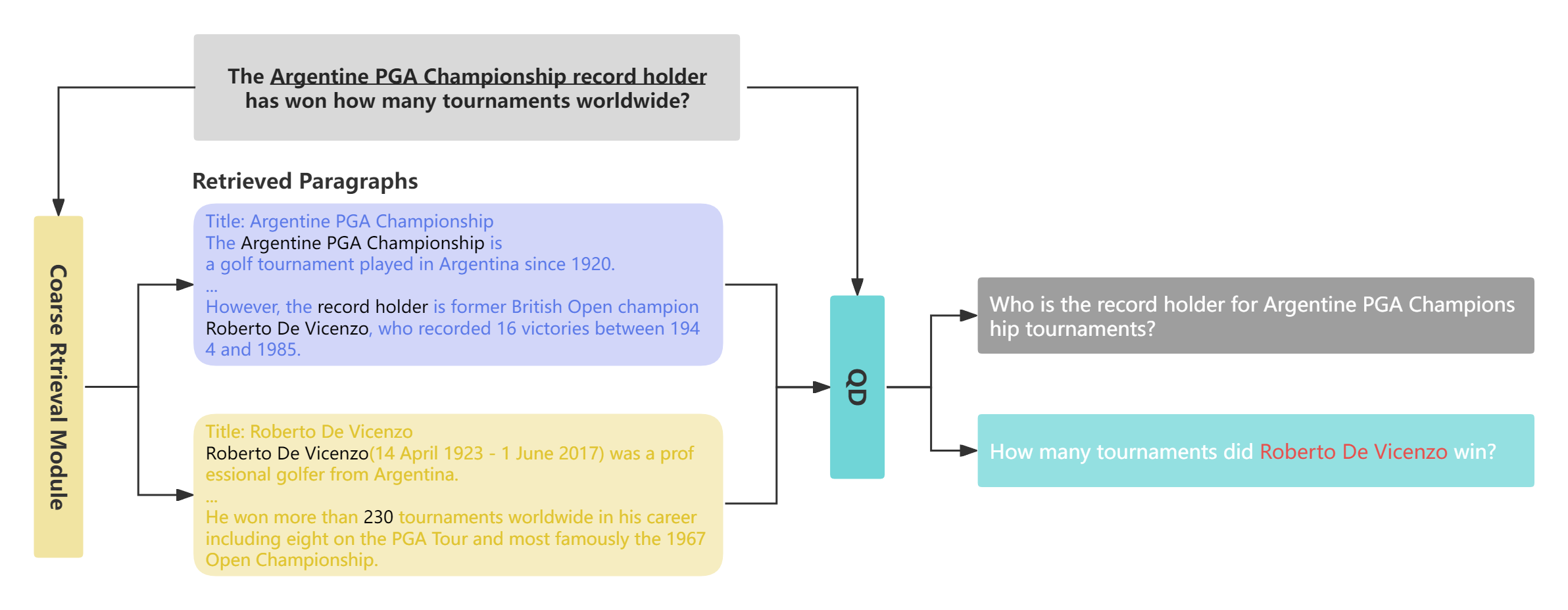}
\caption{\textcolor{black}{Example of multi-hop and decomposed sub-questions from the HotpotQA dataset. The original question is shown in light grey and the decomposed ones are in deep gray and cyan. \emph{"Roberto de Vincenzo"} in the retrieved paragraph is the answer to sub-question Q1 and also part of sub-question Q2. The literal \emph{"230"} is the answer to sub-question Q2. Since the paragraphs are too long, we here only list the sentences that contain supporting facts.}}
\centering
\label{QD_EXAMPLE}
\end{figure*}

Thus, understanding and potentially decomposing multi-hop questions into finer-grained sub-questions is a key desired step in QA. To accurately answer a multi-hop question, traditionally QD + QA methods start by decomposing the given multi-hop question into simpler sub-questions, attempting to answer them in a specific order, and then finally aggregating the information obtained from all sub-questions.

The elucidation and subsequent decomposition of multi-hop questions into more granular sub-questions represent a pivotal step in the domain of Question Answering (QA). Conventional methodologies, namely Question Decomposition (QD) plus QA, commence by segmenting the multi-hop question into simpler, constituent sub-questions. This process involves answering these sub-questions in a predetermined sequence, culminating in the synthesis of information derived from all sub-questions to formulate the final answer.

Our initial exploration reveals that QD constitutes a significant impediment within Multi-hop Question Answering (MHQA). Prior approaches to QD, as documented by \citet{Min2019MultihopRC, Perez2020UnsupervisedQD}, typically bifurcate multi-hop questions into \textbf{dependent} sub-questions. For instance, as illustrated in figure \ref{QD_EXAMPLE}, a multi-hop question is divided into \emph{"Who is the record holder for Argentine PGA Championship tournaments?"} followed by \emph{"How many tournaments did [Answer of Sub Q1] win?"}. This necessitates that QA models accurately resolve the first sub-question and utilize its answer to address the subsequent one, aiming to ascertain the ultimate response. This QD+QA paradigm is prone to error propagation, where inaccuracies in resolving any sub-question can misguide the final answer determination.
Recent work on solving complex or multihop questions is to leverage LLMs with Chain-of-Thought (CoT) \cite{Fu2022ComplexityBasedPF, Wei2022ChainOT} and In-context-learning (ICT) \cite{Liu2021WhatMG} to iteratively decompose and answer complex questions in step-wise, which suffer the error propagation as well. \citet{Venktesh2023InContextAT} hypothesizes that existing specialized QA datasets with rationales or decompositions might already contain instances that can be reused as demonstrations for question decompositions but fail for generalization on other complex QA scenarios.

Our proposed GenDec model alleviates these issues by ensuring that the decomposed sub-questions are independent and self-contained, eliminating the need for sequential answering inherent in previous models. GenDec integrates these sub-questions into the QA model to facilitate an appropriate reasoning pathway.

We introduce \textbf{GenDec}, a generative approach to QD that leverages retrieved paragraphs containing evidential support for segmenting multi-hop questions into independent sub-questions, which do not necessitate ordered answering. Post-QD, GenDec amalgamates the attributes of these sub-questions with mechanisms for relevant paragraph retrieval, supporting facts prediction, and the QA process. As depicted in figure \ref{QD_EXAMPLE}, GenDec's decomposition of a question from the HotpotQA demonstrates this methodology. The original multi-hop inquiry \emph{"The Argentine PGA Championship record holder has won how many tournaments worldwide?"} is segmented into independent sub-questions: \emph{"Who is the record holder for Argentine PGA Championship tournaments?"} and \emph{"How many tournaments did Roberto De Vicenzo win?"}.

GenDec distinguishes itself in the realm of sub-question answering within MHQA tasks by its reliance solely on retrieved paragraphs for decomposing questions, thereby obviating the need to account for the sequence or relational hierarchy of sub-questions. This feature not only simplifies the decomposition process but also enhances the robustness of the model. To underscore the continued significance of Question Decomposition (QD) in the era of Large Language Models (LLMs), we undertake a comprehensive evaluation of GenDec's performance and its generalizability across various MHQA contexts.

Our evaluation framework consists of two pivotal experiments designed to benchmark the performance of the GenDec with a fine-tuned QA model against state-of-the-art QA models. The initial experiment focuses on evaluating various SOTA models in terms of their paragraph retrieval and QA capabilities. Subsequently, the second experiment aims to examine the reasoning and answering prowess of advanced LLMs, including GPT-4 \cite{achiam2023gpt}, GPT-3.5 \cite{Ouyang2022TrainingLM}, and text-davinci-003, particularly when navigating MHQA tasks with the aid of sub-questions. The findings from these experiments collectively affirm that GenDec not only enhances QA performance but also significantly improves paragraph retrieval outcomes across both fine-tuned models and LLMs.

This comprehensive examination and the resultant insights not only highlight GenDec's superior performance and versatility but also reiterate the indispensable role of question decomposition in refining the reasoning abilities of LLMs. The contributions of our work are manifold and can be summarized as follows:
1) We introduce GenDec, an innovative and robust approach that adeptly generates natural language sub-questions leveraging retrieved paragraphs, thereby concealing the underlying reasoning chains. This methodology facilitates a more intuitive and efficient process for question decomposition.
2) Through rigorous experimentation, we demonstrate that GenDec's integration of generated sub-questions into paragraph retrieval and QA modules not only surpasses the performance of existing QD-based QA models but also establishes new benchmarks when compared to other formidable baselines.
3) Our analysis extends to the synergistic combination of GenDec with LLMs, revealing the pivotal role of QD in augmenting the reasoning capabilities of LLMs.

\section{Related Work}
\subsection{Multi-hop Question-answering}
Multi-hop QA requires more than one reasoning step in multiple paragraphs to answer a question. For example, multi-hop QA in DROP  \cite{Dua2019DROPAR} requires numerical reasoning such as addition and subtraction. \citet{yang2018hotpotqa} proposed the HotpotQA dataset that contains 113K multi-hop QA pairs collected from Wikipedia articles by crowd-sourcing. \citet{ho-etal-2020-constructing} presented 2WikiMultiHopQA, which uses structured and unstructured data and introduces the evidence information containing a reasoning path for multi-hop questions. 

\begin{figure*}[ht]
\centering
\includegraphics[width=1.0\textwidth]{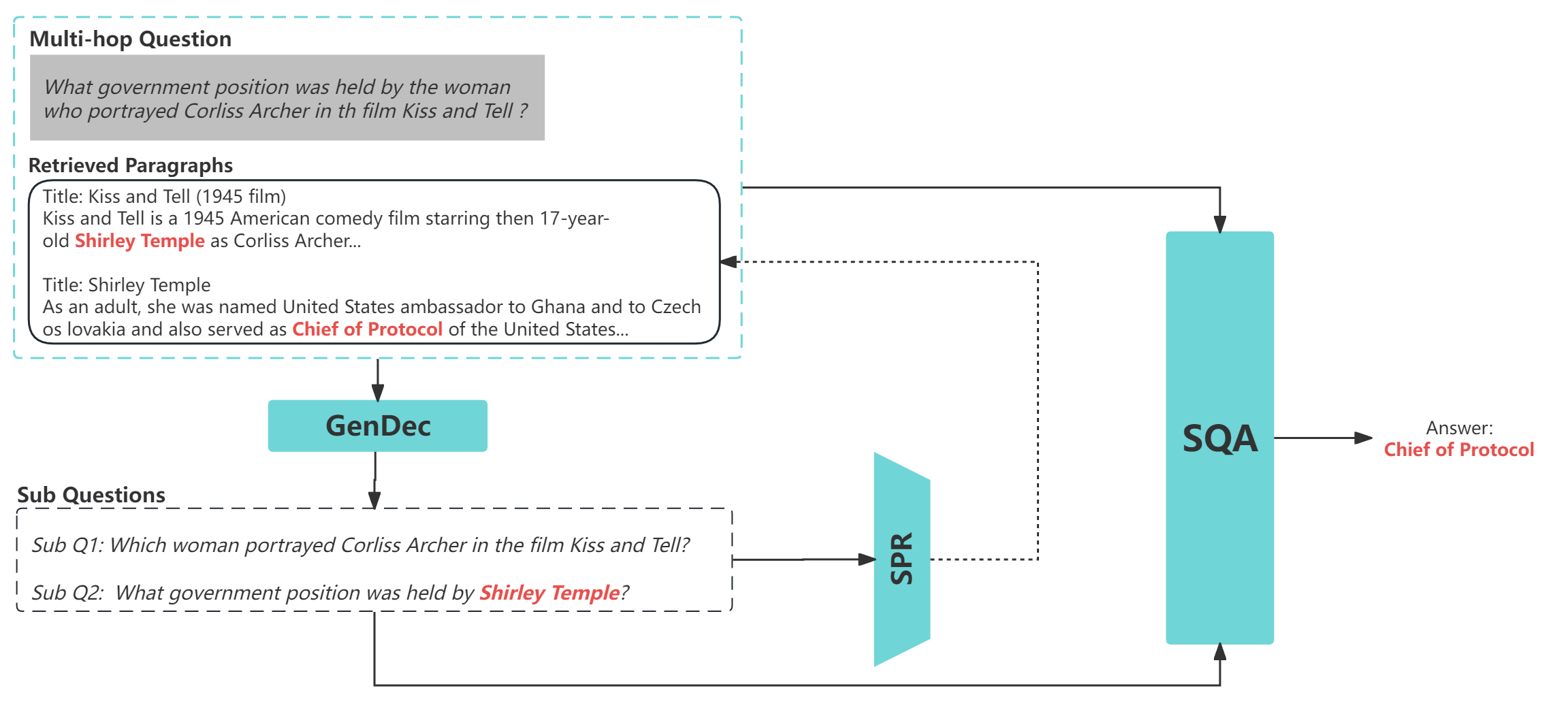}
\caption{ Pipeline of GenDec. From top to bottom. We first carry out Question Decomposition (QD) to decompose a multi-hop question into its sub-questions and then train a \textbf{S}ub-question-enhanced \textbf{P}aragraph \textbf{R}etrieval module (SPR). We then input multi-hop questions, sub-questions, as well as retrieved paragraphs, into the sub-question-enhanced QA module to extract the final answers. }
\centering
\label{Figure framework}
\end{figure*}

\subsection{Question Decomposition}
Several studies conducted QD in complex QA tasks by using different methods. \citet{Wolfson2020Break} and \citet{talmor18compwebq}, inspired by SQL and SPARQL query, proposed rule-based methods. However, they failed to generalize into different types of questions because of the limited rules. \citet{Min2019MultihopRC} proposed a supervised QD method with human-labeling data to predict the text span of sub-questions. ONUS \cite{Perez2020UnsupervisedQD} is a one-to-N unsupervised sequence transduction method that uses supervision information of pseudo-decompositions from Common Crawl to map complex questions into simpler questions and recompose intermediate answers of sub-questions for reasoning final answers. These supervised and unsupervised QD methods decompose complex questions into two sub-questions but are not applicable to real scenarios. \citet{Deng2022InterpretableAQ} trains an Abstract Meaning Representation (AMR)-to-text generation model on the QDMR \cite{Wolfson2020BreakID} dataset. The entity description graph (EDG)-based QD method \cite{Hu2021EDGBasedQD} represents the structure of complex questions to solve the question-understanding and component-linking problems of knowledge base QA tasks. \citet{Zhou2022LearningTD} pre-trained Decomp-T5 on human-collected parallel news to improve the ability of semantic understanding for QD. Instead of answering sub-questions one by one, \citet{Guo2022ComplexRC} directly concatenated sub-questions with the original question and context to leverage the reading-comprehension model to predict the answer. 
\citet{wang2022locate} propose a step-by-step sub-question generation that generates sub-questions at each intermediate step. However, such step-wise reasoning and generation methods suffer from error propagation, while ours can directly generate the sub-questions and reasoning at the same time.

\subsection{LLM reasoning}
LLMs have shown reasoning abilities over several tasks, such as multi-hop QA \cite{bang2023multitask}, commonsense reasoning \cite{Liu2022RainierRK}, and table QA \cite{Chen2022LargeLM}. Chain-of-thought (CoT) \cite{Wei2022ChainOT} leverages a series of intermediate reasoning steps, achieving better reasoning performance on complex tasks. \citet{jin2023tab} proposed a framework called Tabular Chain of Thought (Tab-CoT) that can perform step-by-step reasoning on complex tableQA tasks by creating a table without fine-tuning by combining the table header with related column names as a prompt. \citet{Khot2022DecomposedPA} proposed an approach called Decomposed Prompting to solve complex tasks by decomposing them into simple sub-tasks that can be delegated to a shared library of prompting-based LLMs dedicated to these sub-tasks.

However, these studies only decomposed questions into sub-questions and the latter sub-questions always rely on previous sub-questions. When the previous sub-questions are incorrectly answered, the latter sub-questions are also prone to be misguided.

\section{Framework}
As discussed in the preceding section, previous QD-based QA methods fail to mitigate the error-propagation problem during the answer reasoning process as they decompose questions into sub-questions.
Our framework consists of three main components: (1) a generative QD module, GenDec, to generate independent sub-questions (2) a sub-question-enhanced paragraph-retrieval module, that serves both the supporting facts prediction and QA tasks; and (3) a sub-question enhanced QA module, which fuses features of sub-questions for QA and supporting-facts prediction. Figure \ref{Figure framework} shows the overall framework. 


\subsection{GenDec}
We explore different model architectures for the GenDec, i.e., generative language models (e.g., BART \cite{Lewis2019BARTDS}, T5\cite{2020t5}) as the QD module.

We first leverage the coarse retrieval module proposed by \citet{yin2023rethinking} 
to retrieve relevant paragraphs to serve the GenDec module. In the coarse retrieval module, each question $Q$ is typically combined by a set of $N$ paragraphs ${P_1, P_2\dots, P_N}$, but only a small number of paragraphs (e.g., two in HotpotQA) are labeled as relevant to the question $Q$. We model paragraph retrieval as a binary classification task. Specifically, for each question-paragraph pair, we concatenate it as ``[CLS], question, [SEP], paragraph, [SEP]'' in sequence.

\textbf{Generative Question Decomposition}
To ensure the sub-questions are answerable by the QA module, we train a text-to-text generation model on the sub-question dataset from HotpotQA~\citet{Khot2021TextMN}. Moreover, to improve the generalization of the GenDec module for different types of multihop questions, we also utilize the PokeMQA dataset\cite{Gu2023PokeMQAPK} as part of the training data for the QD module. PokeMQA is an MHQA dataset with corresponding sub-questions (including {2, 3, 4}-hop questions).

We use BART-large and T5-large models as backend models and fine-tune them on the sub-question datasets to generate sub-questions. We use the retrieved paragraphs that contain supporting facts $p$ and question $q$ as input to train a question-generator model $G: (p, q) \Rightarrow sub\_qs$, where $sub\_qs$ is the generated sub-question set. Such a generator, $G$, produces the two sub-questions in the example in Figure \ref{QD_EXAMPLE}. The details of finetuning T5-large and BART-large are given in Appendix \ref{sec:ID}.


\subsection{Sub-question-enhanced Paragraph Retrieval (SPR)}
Multi-hop question answering takes textual context into account and usually, MHQA datasets include multiple paragraphs as question context (e.g., HotpotQA and 2WikiMultiHopQA datasets include 10 paragraphs per question). However, including all such paragraphs is not ideal due to noise and size (length). Therefore, paragraph retrieval plays a vital role in both QA and QD modules, since GenDec utilizes information from sub-questions and can thus focus on the more relevant data.

We propose sub-question-enhanced paragraph retrieval (\textbf{SPR}) for refined retrieval, which utilizes an encoder and a classification head to compute scores for each paragraph to help the supporting facts prediction. Given a $k$-hop question $Q$, generated $k$ sub-questions $q_1,...q_{k}$, and a candidate set with $n$ passages as $\mathcal{P} = {\{p_1, p_2, ..., p_n}\}$, SPR aims to retrieve a relevant paragraph set ($\hat{p}_1, \hat{p}_2, ..., \hat{p}_k$) that relates to the $k$ sub-questions and the $k$-hop question $Q$. While most previous work formulates it as a one- or two-step sequence labeling task, classifying every passage $p_i \in \mathcal{P}$ as relevant or not.

A passage $p_i \in \mathcal{P}$ corresponds to the question $Q$ and j-th sub-question $q_j \in \mathcal{S}$. Consequently, we also denote the output score of SPR as $S(\hat{p}_{i} | Q, q_j)$, given the concatenated sequence of question, sub-question, and passages identified so far, ($Q, q_j, \hat{p}_i$). 

We use the DeBERTa model \cite{he2020DeBERTa} as an encoder to derive embeddings for the concatenated sequence ($Q, q_j, \hat{p}_i$) and the output $\acute{o}_i \in {\mathbb{R}} ^ {n}$. Subsequently, a fully connected layer is added after DeBERTa to project the final dimension of the ``[CLS]'' representations of these embeddings into a 2-dimensional space, representing ``irrelevant'' and ``relevant'' respectively. The logit in the ``relevant'' side serves as the score for each paragraph. This scoring process is denoted by a function $S(\hat{p}_{i} | Q, q_j)$.
In SPR, we optimize the classification of each combination of question, sub-question, and paragraph using Cross-Entropy loss.

\begin{equation}
\begin{aligned}
\mathcal{L}_{{j}}= & - \sum_{q_{i} \in \mathcal{S}} \sum_{\hat{p}_{i} \in \mathcal{P}} l_{j,p} logS(\hat{p}_{i} | Q, q_j) + \\& (1 - l_{j,p})log(1-S(\hat{p}_{i} | Q, q_j))
\end{aligned}
\end{equation}
where $l_{j,p}$ is the label of $\hat{p}_i$ and $S(\hat{p}_i | Q, q_j)$ is the score function predicted by the model.

Thus, we train a paragraph retrieval model based on DeBERTa \cite{he2020DeBERTa} to execute binary classification and rank the scores of paragraphs containing the gold supporting facts.

\subsection{Sub-question-enhanced QA (SQA)}
In the QA task, we propose a sub-question-enhanced QA model which utilizes multi-task learning to simultaneously predict supporting facts, and extract answer spans by incorporating sub-questions produced from GenDec. In order to better evaluate the role of sub-question incorporation, we do not include other additional modules in our model. Instead, we focus on the effects of sub-question incorporation on the performance of SQA. Additionally, as both HotpotQA and 2WikiMultiHopQA datasets also contain questions with yes/no answers, a common scenario, we include an answer type task.

The SQA first combines all retrieved paragraphs into context $C$, which is concatenated with question $Q$ and sub-questions $\{Sub\_Qs\}$ and fed into DeBERTa. We denote the encoded question 
and sub-question representations as $\mathbf{Q}=\{\mathbf{q}_0, \mathbf{q}_1, \ldots, \mathbf{q}_{Q-1}\} \in \mathbf{R}^{m \times d}$ and the encoded context representation as $\mathbf{C}=\{\mathbf{c}_0, \mathbf{c}_1, ..., \mathbf{c}_{C-1}\} \in \mathbf{R}^{C \times d}$, where $Q$ is the length of the question. Each $\mathbf{q}_i$ and $\mathbf{c}_j \in\mathbf{R}^{d}$. 
    \begin{align}
    \nonumber
    \mathbf{P}^i &= \text{DeBERTa} \left(P^{(i)} \right) \\
    \nonumber
    \mathbf{sub\_q}^i &= \text{DeBERTa}\left(Sub\_Q^{(i)} \right) \\
    \mathbf{q} &= \text{DeBERTa}(\mathbf{Q})\,,
    \end{align}
where $P^{(i)}\in \mathbf{R}^{d}$, $Sub\_Q^{(i)} \in \mathbf{R}^{d}$, $\mathbf{Q} \in \mathbf{R}^{d}$ respectively denote the $i$-th paragraph, sub-question, and question representations. 

To extract answer spans, we use a linear prediction layer on the contextual representation to identify the start and end positions of answers and employ cross-entropy as the loss function. The corresponding loss terms are denoted as $\mathcal{L}_{start}$ and $\mathcal{L}_{end}$, respectively. The classification loss for the supporting facts is denoted as $\mathcal{L}_{sup}$, and we jointly optimize all of these objectives in our model. We also introduce an answer type classification module trained with crossentropy loss function.
\begin{equation}
\label{eq:answer_type_loss}
    \mathcal{L}_{type}=\mathbb{E}[-\sum_{i=1}^{3}y_{i}^{type}log(\hat{y}_{i}^{type})]
\end{equation}
where $\hat{y}{i}^{fine}$ denotes the predicted probability of answer types classified by our model, and $y{i}^{fine}$ represents the corresponding one-hot encoded ground-truth distribution. $y_{i}^{type}$ has three values: 0 denotes a negative answer, 1 denotes a positive answer, and 2 denotes the answer is a span.

The multi-task prediction model's total loss is:
\begin{equation}
\label{eq:reading_loss}
    \mathcal{L}_{reading}= \lambda_1 \mathcal{L}_{type}+\lambda_2 (\mathcal{L}_{start}+\mathcal{L}_{end})+\lambda_3\mathcal{L}_{sup}
\end{equation}

Similarly, we set $\lambda_1$, $\lambda_2$, and $\lambda_3$ all to 1, giving equal importance to each module for multitask learning. The implementation details of the Sub-question-enhanced QA module are described in Appendix \ref{sec:ID}.

\section{Experiments and Analysis}
This section describes the utilized datasets and we first illustrate the performance of SQA in the QA task in section \ref{SQA}; then we show the impact of GenDec with LLMs in section \ref{GenDec+LLM}; To describe the GenDec effectiveness on paragraph retrieval, we conduct and analyze experiment results in section \ref{SPR}; To further show the effectiveness of SQA on reasoning chain, we list results in section \ref{RC}; We analyze the ablation study of GenDec on SQA in section \ref{ablation}. We also illustrate the quality of generated sub-questions to describe the effectiveness, generalization and robustness of GenDec in appendix \ref{Quality analysis}.
\subsection{Datasets}

\textbf{Question Answering (QA)} We evaluate GenDec on the MuSiQue~\cite{trivedi2021musique}, 2WikiMultiHopQA~\cite{xanh2020_2wikimultihop} and HotpotQA~\cite{yang-etal-2018-hotpotqa} datasets, which contain 20K, 160K and 90K training instances. These three multi-hop QA datasets consist of questions, answers, supporting facts, and a collection of 10 paragraphs as context per question.\\

\textbf{Question Decomposition (QD)} To train and evaluate GenDec, we use the sub-questions and answers data processed from the multi-hop HotpotQA dataset~\citet{Khot2021TextMN} - here named SQA for clarity. These sub-questions are relatively high quality, in that we are able to use them to train a sub-question generator that achieves high task performance on multi-hop QD. 
However, most of the questions in the HotpotQA dataset are 2-hop questions, to improve the generalization of GenDec, we here add the PokeMQA \cite{Gu2023PokeMQAPK} (including {2, 3, 4}-hop questions) as the data augmentation. PokeMQA contains 3000 multi-hop questions (1000 2-hop questions, 1000 3-hop questions, and 1000 4-hop questions) with corresponding sub-questions. We divided the PokeMQA into the train and test sets, and the split ratio is 80\% for the train set and 20\% for the test set.
It is important to note that once trained, the GenDec module can be re-used across scenarios, without depending on dataset-specific decomposed questions of golden annotations.\\

\textbf{Sub-question Reasoning} To evaluate the reasoning ability of GenDec with LLMs and finetuned QA models, we also utilize a human-verified sub-question test dataset derived from HotpotQA~\citet{tang2020multi} - here named HVSQA for clarity; which provides a strong benchmark to evaluate QA models in answering complex questions via sub-question reasoning.

\subsection{Experiment Results}
\begin{table*}[t!]
\centering
\begin{adjustbox}{scale=0.8,center}
\begin{tabular}{lcccccc}
\toprule
\multirow{2}{*}{Model} & \multicolumn{2}{c}{Ans} & \multicolumn{2}{c}{Sup} & \multicolumn{2}{c}{Joint} \\ 
& EM & F1 & EM & F1 & EM & F1 \\ \midrule
\rowcolor{gray!10} \multicolumn{7}{c}{\textbf{\textsl{HotpotQA}}} \\ \hline
\multicolumn{7}{c}{QD-based QA Models} \\
DecompRC \cite{Min2019MultihopRC} & 55.20 & 69.63 & - & - & - & - \\ 
ONUS \cite{Perez2020UnsupervisedQD} & 66.33	& 79.34 & - & - & - & - \\
\midrule
\multicolumn{7}{c}{GNN-based Models} \\
DFGN \cite{Xiao2019DynamicallyFG} & 56.31 & 69.69 & 51.50 & 81.62 & 33.62 & 59.82 \\
SAE-large \cite{tu2020select} & 66.92 & 79.62 & 61.53 & 86.86 & 45.36 & 71.45 \\
C2F Reader\cite{Shao2020IsGS} & 67.98 & 81.24 & 60.81 & 87.63 & 44.67 & 72.73 \\
HGN-large \cite{Fang2019HierarchicalGN} & 69.22 & 82.19 & 62.76 & 88.47 & 47.11
& 74.21 \\
BRF-graph \cite{huang2021breadth} & 70.06 & 82.20 & 61.33 & 88.41 & 45.92 & 74.13 \\
AMGN+ \cite{ijcai2021p531}  & 70.53 & 83.37 & 63.57 & 88.83 & 47.77 & 75.24 \\
\midrule
\multicolumn{7}{c}{Other STATE-OF-The-ART(SOTA) Models} \\
FE2H on ALBERT \cite{Li2022FromET} & 71.89 & 84.44 & 64.98 & 89.14 & 50.04 & 76.54 \\
PCL \cite{deng2022prompt} & 71.76 & 84.39 & 64.61 & 89.20 & 49.27 & 76.56\\
Smoothing R3 \cite{yin2023rethinking} & 72.07 & 84.34 & 65.44 & 89.55 & 49.73 & 76.69\\
Beam Retrieval \cite{zhang2023beam} & \textbf{72.69}	& \textbf{85.04} &	\textbf{66.25} &	\underline{90.09} &	\textbf{50.53} &	\textbf{77.54} \\
\midrule
SQA (ours) & \underline{72.39} & \underline{84.69} & \underline{65.88} & \textbf{90.31} & \underline{50.34} & \underline{77.48} \\
\midrule
\rowcolor{gray!10} \multicolumn{7}{c}{\textbf{\textsl{2WikiMultihopQA}}} \\ \hline
 CRERC \cite{CRERC} & 69.58 & 72.33 & 82.86 & 90.68 & 49.80 & 58.99\\
NA-Reviewer \cite{NA-Reviewer} & 76.73 & 81.91 &  89.61 & 94.31 & 52.75 & 65.23 \\
BigBird-base model \cite{ho-etal-2023-analyzing} & 74.05 & 79.68 & 77.14 & 92.13 &39.30	& 63.24 \\
Beam Retrieval \cite{zhang2023beam} & \textbf{88.47} & \textbf{90.87} & \textbf{95.87} & \textbf{98.15} &	- & - \\
SQA (ours)& \underline{86.47} & \underline{88.15} & \underline{93.28} & \underline{96.45} & \underline{56.87} & \underline{68.38} \\
\midrule
\rowcolor{gray!10} \multicolumn{7}{c}{\textbf{\textsl{MuSiQue-Ans}}} \\ \hline
Beam Retrieval (beam size 2) \cite{zhang2023beam} & - & \textbf{69.20} & - & \textbf{91.40} & - & -\\
Beam Retrieval (beam size 1) \cite{zhang2023beam} & - & 66.90 & - & 90.00 & - & - \\
Ex(SA) \cite{trivedi2021musique} & - & 49.00 & - & 78.10 & - & - \\
Ex(EE) \cite{trivedi2021musique} & - & 46.40 & - & 80.60 & - & - \\
SQA (ours)& - & \underline{65.40} & - & \underline{87.90} & - & - \\
\hline
\bottomrule
\end{tabular}
\end{adjustbox}
\caption{\label{tab:leaderboard_distractor} Performance of different QA models on test distractor settings of HotpotQA, 2WikiMultihopQA and MuSiQue Answerable datasets. GenDec outperforms all QD-based and other GNN-based QA models.}
\end{table*}

\begin{table}[t]
\centering
\begin{adjustbox}{scale=0.85,center}
\begin{tabular}{lccccl}
\hline
\textbf{Model} & \textbf{EM} & \textbf{F1} \\
\hline
 $\text{SAE}_{large}$~\cite{tu2020select}& 91.98 & 95.76 \\
 $\text{S2G}_{large}$~\cite{wu2021graph}& 95.77 & 97.82 \\
 $\text{FE2H}_{large}$~\cite{li2022easy}& 96.32 & 98.02 \\
 $\text{C2FM}_{large}$~\cite{yin2023rethinking}& 96.85 & 98.32 \\
 $\text{Beam Retrieval}_{large}$~\cite{zhang2023beam}& 97.52 & 98.68 \\
 \hline
 SPR (ours) & 97.53 & 98.78 \\
 GenDec + Beam Retrieval & \textbf{98.02} & \textbf{99.17} \\
 \hline
\end{tabular}
\end{adjustbox}
\caption{\label{tab:PR_results}
Comparison of our SPR with previous baselines on HotpotQA dev set.}
\end{table}

\begin{table}[t]
\begin{adjustbox}{scale=0.85,center}
	\begin{tabular} {m{0.1cm}<{\centering} m{0.5cm}<{\centering} m{0.6cm}<{\centering}| m{0.7cm}<{\centering} m{0.8cm}<{\centering} m{0.5cm}<{\centering} m{0.5cm}<{\centering} m{0.5cm}<{\centering} m{1.1cm}<{\centering}} \cline{1-9}
		$q$ &$q_{sub1}$ &$q_{sub2}$ &DFGN &DecRC &HGN &PCL & BR & SQA \\ \cline{1-9}	
		\cellcolor{green!40}c 	&\cellcolor{green!40}c	&\cellcolor{green!40}c  &23    &31.3	&39.5  &43.6   & 52.2 & 52.9  \\ \cline{1-9}			
            c 	&c	&w	&9.7   &7.2  	&5.1   &6.8    & 7.8  & 6.4   \\ \cline{1-9}		
           \cellcolor{red!40} c 	& \cellcolor{red!40}w	& \cellcolor{red!40}c  &17.9  &19.1	&19.6  &21.3   & 21.7 & 18.8  \\ \cline{1-9}		
            c 	&w	&w  &7.5   &5.5 	&3.8   &2.1    & 6.7  & 6.2   \\ \cline{1-9}			
            \cellcolor{red!40}w	&\cellcolor{red!40}c	&\cellcolor{red!40}c  &4.9   &3.0	    &2.8   &1.7    & 1.2  & 1.1   \\ \cline{1-9}			
            w	&c	&w  &17	   &18.6	&16.7  &16.3   & 7.3  & 10.3  \\ \cline{1-9}		
            \cellcolor{red!40}w	&\cellcolor{red!40}w	&\cellcolor{red!40}c  &3.5   &3.4     &2.6   &1.1    & 0.7  & 0.9   \\ \cline{1-9}		
            w	&w	&w  &16.5  &11.9	&9.9   &7.1    & 2.4  & 3.4   \\ \cline{1-9}
	\end{tabular}
	\vspace{-2mm}
 \end{adjustbox}
	\caption{Categorical EM statistics (\%) of sub-question evaluation for six multi-hop QA models over HVSQA \cite{tang2020multi}. c/w denotes questions answered correctly/wrongly. For example, the fourth row shows the percentage of multi-hop questions that can be correctly answered while sub-questions cannot.}
\label{sub-qa}
\end{table}

\subsection{SQA against SOTA QA systems}\label{SQA}
We use Exact Match (EM) and F1 scores as evaluation metrics for answer span prediction and supporting facts prediction on the HotpotQA, 2WikiMultiHopQA, and MuSiQue-Ans datasets to compare the performance of SQA with that of QD-based, GNN-based, and other SOTA QA models. 

As shown in Table \ref{tab:leaderboard_distractor}, SQA With the Deberta-large model outperforms most models in both metrics, including the strong baseline, and performs very competitively to the latest SOTA on the HotpotQA dataset, only below the Beam retrieval \cite{zhang2023beam}.

The middle and bottom sections of the table also show that SQA significantly outperforms most previous work on the 2WikiMultiHopQA and MuSiQue-Ans \cite{trivedi2021musique} datasets. GenDec's performance is only lower than the contemporary Beam Retrieval \citet{zhang2023beam}, which takes a retrieval approach that can be complementary to SQA itself.

\subsection{GenDec with LLMs}\label{GenDec+LLM}
We also evaluate the impact of generated independent sub-questions on LLM reasoning in table \ref{tab:LLM}. 
We also evaluated the performance of LLMs with and without GenDec on 1000 samples of dev distractor settings. Figure \ref{Figure prompt} shows the used with QD and without QD prompt settings. We selected the 1-shot setting in which LLMs are given one example from the training set with two prompts, one is reasoning over sub-questions and the other is directly reasoning answers. As shown in Table \ref{tab:LLM}, GPT-4 with additional sub-question information performs better than without sub-questions. GPT-4 with QD prompting achieves higher answer span extraction on the F1 score (76.28) and EM (56.24) respectively. 

Moreover, we also evaluate the reasoning chain of LLMs with and without GenDec, we here follow the experiment settings of \citet{tang2020multi} and calculate the proportion of reasoning chains which is shown in table \ref{LLM_reasoning_chain} in the appendix. For instance, GPT-4 gets 39.7\% right reasoning chain (green row) which achieves a correct right final answer based on right sub-answer1 and sub-answer2. However, the wrong reasoning chain
is 21.1\% (sum of the three red rows) although a right final answer either sub-answer1 or sub-answer2 is wrong. When removing the GenDec module, GPT-4 only gets a 30.5\% right reasoning chain, and 24.5\% wrong reasoning chain, deeming that GenDec generated parallel sub-questions could help LLMs achieve a higher performance following a higher proportion of reasoning chain.

\subsection{SPR on paragraph retrieval}\label{SPR}
Table \ref{tab:PR_results} shows the SOTA paragraph retrieval performance of GenDec's SPR method against previous strong paragraph retrieval model baselines. SPR reaches very competitive results against Beam Retrieval (slightly higher in F1 vs slightly lower EM). Moreover, combining our GenDec approach with Beam Retrieval further improves performance and showcases the efficacy of leveraging sub-questions. GenDec + Beam Retrieval achieves SOTA performance on paragraph retrieval, showing the high quality of retrieved paragraphs.

\subsection{SQA on reasoning chain evaluation}\label{RC}
To analyze whether existing multi-hop QA models can demonstrate the right reasoning process, we compare the percentage of correct final answers and intermediate answers obtained by DFGN \cite{Xiao2019DynamicallyFG}, DecompRC (DecRC) \cite{Min2019MultihopRC}, HGN \cite{Fang2019HierarchicalGN}, PCL \cite{deng2022prompt}, Beam Retrieval (BR) \cite{zhang2023beam}, and our SQA.
Table \ref{sub-qa} summarizes the percentage of correct answers for intermediary sub-questions and final multi-hop questions on the HVSQA \cite{tang2020multi} dataset. We observe that, although Beam Retrieval (BR) achieves a relatively high percentage of correct final answers (88.3\% vs SQA's 84.3\%), 36.2\% of these are obtained from incorrect intermediate answers and reasoning chains. While SQA's performance is less affected by this case, at 31.4\%. Comparing fully correct answer chains, Beam Retrieval and SQA reach 52.2\%, and 52.9\% respectively. The experiment results reveal that previous SOTA QA systems are not reliable enough and get an inflated performance as they usually bypass the right reasoning chain.

\subsection{Ablation Study}\label{ablation}
\begin{table}[t]
\small
\centering
\begin{adjustbox}{scale=0.85,center}
\begin{tabular}{ccccccc}
\toprule
\multirow{2}{*}{Model} & \multicolumn{2}{c}{Q\_ori} & \multicolumn{2}{c}{Q\_sub1} & \multicolumn{2}{c}{Q\_sub2} \\ \cmidrule{2-7}
& F1 & EM & F1 & EM & F1 & EM \\ 
\midrule
CogQA & 67.82 & 53.2 & 69.65 & 58.6 & 68.49 & 54 \\ 
DFGN & 71.96 & 58.1 & 68.54 & 54.6 & 60.83 & 49.3 \\ 
DecompRC & 77.61 & 63.1 & 75.21 & 61 & 70.77 & 56.8 \\ 
ONUS & 79.25 & 67.43 & 77.56 & 63.89 & 72.21 & 57.62 \\ 
PCL & 73.8 & 87.15 & 68.4 & 83.62 & 68.5 & 81.07 \\
\midrule
SQA w/o GenDec & 82.81 & 70.72 & 87.45 & 72.65 & 80.12 & 70.38 \\ 
SQA with GenDec & \textbf{86.17} & \textbf{72.88} & \textbf{90.52} & \textbf{76.43} & \textbf{84.61} & \textbf{74.83} \\ 
\bottomrule
\end{tabular}
\end{adjustbox}
\caption{Performance comparison between SQA (with and without the GenDec module) and other QA models on HVSQA~\cite{tang2020multi}, a human-verified sub-question test dataset from HotpotQA.}
\label{reasoning_ablitity}
\end{table}
To evaluate the impact of the GenDec module on the SQA model, we conduct an ablation study testing the performance of answering all sub-questions and original questions, with and without the GenDec module. The results, shown in Table \ref{reasoning_ablitity}, indicate that the GenDec module shows consistent and significantly improved results; improving the F1 score and EM by 3.36 and 2.16, respectively, in the original QA. In answering intermediate answers to sub-questions, SQA w/ GenDec also improves over w/o GenDec (improving the F1 score and EM by 3.07 and 3.78, and 4.49 and 4.45 on sub-questions 1 and 2 respectively). The results indicate that the GenDec module plays an important role in QA tasks not only in its QA ability but also in high performance of intermediate answer reasoning to support answering the final question. 

\section{Conclusion}
We proposed GenDec, a generative-based QD method that generates complete and independent sub-questions based on incorporating retrieved paragraphs. We train an SPR and an SQA module that incorporates sub-questions generated by GenDec and show that it significantly improves QA and paragraph retrieval tasks. We also explore the impact of GenDec on LLMs. Lastly, while GenDec outperforms all previous QD-based and GNN-based QA systems in multi-hop QA, it can still face errors due to incorrect supporting fact predictions influencing the model to incorrectly predict both sub-questions and final answers.

\section{Limitations}
In this paper, we focus on the impact of GenDec in multi-hop QA, where the answers to most questions can be decomposed into several independent sub-questions via the fusion of retrieved paragraphs. Although GenDec performs very well on QD and help improve QA, one of its limitations is that it is still sensitive to errors in paragraph retrieving. The QD results would be affected when incorrect paragraphs are selected. For future work, we plan to focus on tackling this problem.


\bibliography{anthology,custom}
\bibliographystyle{acl_natbib}

\appendix
\section{Implementation Details}
\label{sec:ID}


\begin{table*}[t!]
\centering
\begin{adjustbox}{scale=1.0,center}
\begin{tabular}{lcccccc}
\toprule
\multirow{2}{*}{Model} & \multicolumn{2}{c}{Ans} & \multicolumn{2}{c}{Sup} & \multicolumn{2}{c}{Joint} \\ 
& EM & F1 & EM & F1 & EM & F1 \\ \midrule
GPT-4 w/o GenDec & 61.26 & 78.93 & 64.25 & 85.43 & 48.72 & 75.31 \\
GPT-4 w GenDec & \textbf{64.34} & \textbf{82.46} & \textbf{67.74} & \textbf{88.45} & \textbf{52.43} & \textbf{78.92} \\ \hline
GPT-3.5 w/o GenDec & 57.35 & 75.11 & 59.78 & 76.46 & 42.37 & 69.24  \\
GPT-3.5 w GenDec & \textbf{60.47} & \textbf{77.59} & \textbf{63.25} & \textbf{80.34} & \textbf{47.62} & \textbf{72.27} \\ \hline
text-davinci-003 w/o GenDec & 40.37 & 59.46 & 47.52 & 62.96 & 36.62 & 55.89 \\
text-davinci-003 w GenDec & \textbf{44.39} & \textbf{65.48} & \textbf{50.16} & \textbf{67.28} & \textbf{40.56} & \textbf{61.75} \\
\bottomrule
\end{tabular}
\end{adjustbox}
\caption{\label{tab:LLM} Performance of LLMs (with and without GenDec) on 1000 samples from HotpotQA's dev set distractor setting data.}
\end{table*}

\paragraph{Question Decomposition}
We use the pre-trained T5-large and BART-large models with max\_input\_length $L$ = 512, and max\_output\_length $O$ = 64. During training, we used the Adam optimizer in the QD modules and set the batch size to 32 and the learning rate to 5e-5. All experiments utilized two TITAN RTX GPUs and 5 hours in total. 
\paragraph{Question Answering}
We choose DeBERTa-v2-large as the backend model and set the number of epochs to 12 and batch size to 4. We use BERTAdam with a learning rate of 5e-6 for the optimization and set max position embeddings to 1024. 

\begin{table*}[t]
\begin{adjustbox}{scale=0.85,center}
	\begin{tabular} {m{0.1cm}<{\centering} m{0.5cm}<{\centering} m{0.6cm}<{\centering}| m{2.0cm}<{\centering} m{2.5cm}<{\centering} m{2.5cm}<{\centering} m{3.0cm}<{\centering} m{2.0cm}<{\centering} m{2.0cm}<{\centering}} \cline{1-9}
		$q$ &$q_{sub1}$ &$q_{sub2}$ &GPT-4 w GD & GPT-4 w/o GD &GPT-3.5 w GD & GPT-3.5 w/o GD & text w GD & text w/o GD \\ \cline{1-9}	
		\cellcolor{green!40}c &\cellcolor{green!40}c &\cellcolor{green!40}c & 39.7 & 30.5 & 31.1 & 28.3 & 29.8 & 27.4   \\ \cline{1-9}			
             c 	&c	&w	                                                       & 10.5 & 9.1  & 12.3 & 11.5 & 9.7  & 12.3\\ \cline{1-9}		
            \cellcolor{red!40}c &\cellcolor{red!40}w &\cellcolor{red!40}c       & 12.6 & 9.7  & 8.6  & 9.2  & 8.8  & 11.8  \\ \cline{1-9}		
             c 	&w	&w                                                         & 20.3 & 21.4 & 23.8 & 24.5 & 20.5 & 16.2   \\ \cline{1-9}			
            \cellcolor{red!40}w	&\cellcolor{red!40}c &\cellcolor{red!40}c       & 4.9  & 8.1  & 7.8  & 5.7  & 9.9  & 11.5   \\ \cline{1-9}			
             w	&c	&w                                                         & 6.7  & 7.6  & 5.7  & 6.3  & 6.8  & 7.6  \\ \cline{1-9}		
            \cellcolor{red!40}w	&\cellcolor{red!40}w	&\cellcolor{red!40}c    & 3.6  & 6.7  & 2.6  & 8.4  & 7.9  & 5.8  \\ \cline{1-9}		
             w	&w	&w                                                         & 7.6  & 5.9  & 8.1  & 6.1  & 6.6  & 7.4  \\ \cline{1-9}
	\end{tabular}             
	\vspace{-2mm}
 \end{adjustbox}
	\caption{Categorical EM statistics (\%) of sub-question evaluation for six multi-hop QA models over HVSQA \cite{tang2020multi}. c/w denotes questions answered correctly/wrongly. For example, the fourth row shows the percentage of multi-hop questions that can be correctly answered while sub-questions cannot. We abbreviate GenDec as GD and text-davinci-003 as text.}
\label{LLM_reasoning_chain}
\end{table*}

\begin{table*}
\centering
\small
\begin{tabular}{lcccc}
\toprule
& \multicolumn{3}{c}{\textbf{Metric}}\\
\cmidrule(lr){2-5}
\textbf{Models} & \textbf{F Measure} & \textbf{Rouge1} & \textbf{Rouge-L} & \textbf{BLEU}\\
\midrule
\textsc{BART-large} & \textbf{$76.52$} & \textbf{$75.89$} & \textbf{$64.48$} & \textbf{$28.91$} \\
\textsc{T5-large} & $72.68$ & $74.14$ & $62.22$ & $27.87$\\
\bottomrule
\end{tabular}
\caption{Generative QD performance of different generative LMs on test instances of \textsc{HotPotQA} sub-questions. Results are averaged on 1549 test instances.}
\label{table: QD_quality}
\end{table*}

\section{GenDec Quality Analysis}\label{Quality analysis}
We first quantitatively compare the F measure, ROUGE-1, ROUGE-L, and BLEU scores of the generated sub-question qualities, and BART-large significantly improves the quality of sub-questions reaching 76.52, 75.89,64.48 and 28.91, respectively. 
Likely due to the different max input lengths of T5-large (512) and BART-large (1024), BART outperforms it since some inputs contain more sentences (including both the multi-hop questions themselves and retrieved paragraphs).

We further compare the decomposition quality of GenDec and previous QD models in Table \ref{tab:question_decomposition_examples_for_3_decomposers}, which illustrates how GenDec can produce sub-questions with more natural and fluent language with higher quality. 
Table \ref{cases} further shows some examples of our generated sub-questions.
The GenDec produced sub-questions are also independent that can be answered in parallel. However, the ModularQA failed to generate fluent sentences and DecompRC can not produce independent sub-questions. 
To illustrate the generalization of GenDec on different questions, we list 3 multihop questions (including {2, 3, 4}-hop questions) and generated sub-questions in table \ref{tab:question_decomposition_generalization}, and the results show that our GenDec is robust enough when the complexity of multihop questions improves.

\begin{table*}[t!]
\centering
\small
\begin{tabular}{p{0.90\linewidth}}
\toprule
\textbf{\textsc{GenDec (Ours)}} \\
\midrule[0.03em]
\textbf{Sub-question 1:} Which South Korean boy group had their debut album in 2014? \\
\textbf{Sub-question 2:} WINNER was formed by who? \\


\toprule[0.03em]
\textbf{\textsc{ModularQA} \cite{Khot2021TextMN}} \\
\midrule[0.03em]
\textbf{Sub-question 1:} What is the name of the South Korean group that had their debut album in 2014?\\
\textbf{Sub-question 2:} What was WINNER formed by? \\

\toprule[0.03em]
\textbf{\textsc{DecompRC} \cite{Min2019MultihopRC}} \\
\midrule[0.03em]
\textbf{Sub-question 1:} 2014 S/S is the debut album of which South Korean boy group? \\
\textbf{Sub-question 2:} which formed by who ? \\
\bottomrule
\end{tabular}

\caption{QD examples produced by \{\textsc{GenDec}, \textsc{ModularQA}, \textsc{DecompRC}\} for question ``2014 S/S is the debut album of a South Korean boy group that was formed by who?''.}
\label{tab:question_decomposition_examples_for_3_decomposers}
\end{table*}

\begin{table*}[t!]
\centering
\small
\begin{tabular}{p{0.90\linewidth}}
\toprule
\textbf{\textsc{2 hop Question} } \\
\midrule[0.03em]
\textbf{Original Question} What is the capital city of the country that Michael Feinstein is a citizen of ?\\
\textbf{Sub-question 1:} What is the country of citizenship of Michael Feinstein? \\
\textbf{Sub-question 2:} What is the capital of the United States of America? \\

\toprule[0.03em]
\textbf{\textsc{3 hop Question} } \\
\midrule[0.03em]
\textbf{Original Question} What is the work location of the person who founded the religion of Pius VI?\\
\textbf{Sub-question 1:} Which religion is Pius VI associated with?\\
\textbf{Sub-question 2:} Who founded Catholic Church? \\
\textbf{Sub-question 3:} What is the work location of Jesus Christ? \\

\toprule[0.03em]
\textbf{\textsc{4 hop Question} } \\
\midrule[0.03em]
\textbf{Original Question} What is the capital of the country where Mel Daniels' sport originated from? \\
\textbf{Sub-question 1:} What position does Mel Daniels play? \\
\textbf{Sub-question 2:} Which sport is the center associated with? \\
\textbf{Sub-question 3:} Which country was basketball from?\\
\textbf{Sub-question 4:} What is the capital of the United States of America?\\
\bottomrule
\end{tabular}

\caption{Generalization of \{\textsc{GenDec}\}, examples produced by \{\textsc{GenDec}\} for different questions (including {2, 3, 4}-hop questions).}
\label{tab:question_decomposition_generalization}
\end{table*}

\begin{table*}[t!]
\centering
\small
\begin{tabular}{p{5cm}p{5cm}p{3cm}p{2cm}}
\toprule
Original Question & Sub-questions & Intermediate Answers & Answer\\
\toprule
Were Scott Derrickson and Ed Wood of the same nationality? & What was Scott Derrickson’s nationality? What was Ed Wood’s nationality? \Checkmark & American \Checkmark & Yes \Checkmark \\

\toprule[0.03em]
What government position was held by the woman who portrayed Corliss Archer in the film Kiss and Tell? & Who portrayed Corliss Archer in Kiss and Tell? What position was held by Shirley Temple? \Checkmark & Shirley Temple \Checkmark & Chief of Protocol \Checkmark \\
\midrule[0.03em]
The director of the romantic comedy \"Big Stone Gap\" is based in what New York City neighborhood? & Who is the director of the romantic comedy Big Stone Gap? In what New York City neighborhood is Adriana Trigiani based? \Checkmark & Adriana Trigiani \Checkmark & Greenwich Village \Checkmark \\
\midrule[0.03em]
Are Random House Tower and 888 7th Avenue both used for real estate? & The Random House Tower used as real estate? What is 888 7th Avenue used also for? \XSolidBrush & Used \XSolidBrush & No \XSolidBrush \\

\toprule[0.03em]
What is the name of the executive producer of the film that has a score composed by Jerry Goldsmith? & What is the name of the film of which Jerry Goldsmith composed the score? Which co-writer of Alien was also an executive producer? \Checkmark & Alien \Checkmark & Francis Ford Coppola \XSolidBrush\\
\midrule[0.03em]
Alvaro Mexia had a diplomatic mission with which tribe of indigenous people?& Who was given a diplomatic mission to the native populations living south of St. Augustine and in the Cape Canaveral area? What is the name of the indigenous tribe of Florida? \XSolidBrush & Alvaro Mexia \XSolidBrush& Indigenous peoples of Florida \XSolidBrush\\
\bottomrule
\end{tabular}
\caption{Examples of 3 correct samples and 3 incorrect samples from dev set of HotpotQA}
\label{cases}
\end{table*}

\begin{figure*}[ht]
\centering
\includegraphics[width=1.1\textwidth]{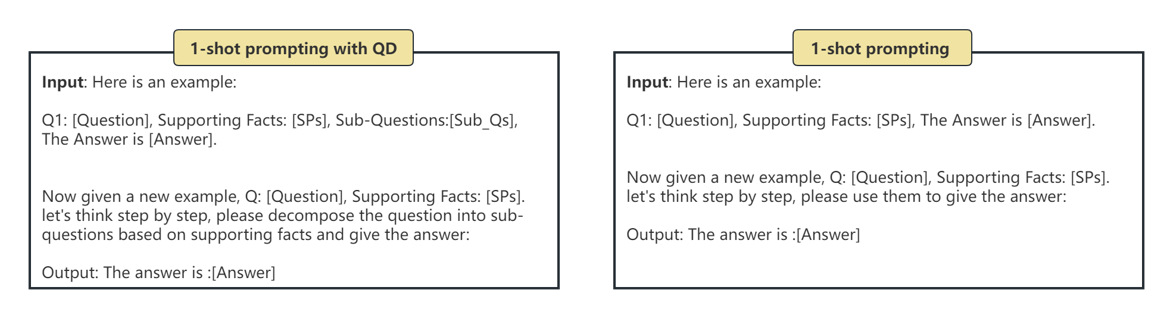}
\caption{Prompting examples of different settings.}
\centering
\label{Figure prompt}
\end{figure*}

\end{document}